\documentclass[runningheads]{llncs}

\usepackage[T1]{fontenc}
\usepackage[utf8]{inputenc}
\usepackage{lmodern}
\usepackage{graphicx}
\usepackage{amsmath,amssymb,amsfonts}
\usepackage{algorithmic}
\usepackage{graphicx}
\usepackage{textcomp}
\usepackage{url}
\usepackage{xcolor}
\usepackage{pifont} 
\usepackage{hyperref}
\usepackage[authoryear]{natbib}

\usepackage{color}

\def\BibTeX{{\rm B\kern-.05em{\sc i\kern-.025em b}\kern-.08em
    T\kern-.1667em\lower.7ex\hbox{E}\kern-.125emX}}
    
\newcommand{\VOCA}{VOCAGEN}
\newcommand{\Tykomz}{Tykomz}

\usepackage[english]{babel}
\usepackage{csquotes}

\begin{document}
\title{Deep Learning and Natural Language Processing in the Field of Construction.}
\author{Rémy Kessler
\and
Nicolas Béchet}
%
\authorrunning{Rémy Kessler et al.}
%
\institute{Université Bretagne-Sud\\
CNRS 6074A\\
56017 Vannes, France\\
\email{nicolas.bechet}@univ-ubs.fr\\  
,\\ 
LIA / Université d’Avignon\\ 
339 chemin des Meinajariès,\\ 
84911 Avignon\\
\email{remy.kessler}@univ-avignon.fr\\ 
}
\titlerunning{Deep L. and NLP in the Field of Construction.}
\maketitle

\begin{keywords}
terminology extraction, Internet queries, linguistic patterns,  hypernym detection, embedding models, and deep learning.
\end{keywords}

\begin{abstract}
 This article presents a complete process to extract hypernym relationships in the field of construction using two main steps: terminology extraction and detection of hypernyms from these terms.
 We first describe the corpus analysis method to extract terminology from a collection of technical specifications in the field of construction. Using statistics and word n-grams analysis, we extract the domain's terminology and then perform pruning steps with linguistic patterns and internet queries to improve the quality of the final terminology.
Second, we present a machine-learning approach based on various words embedding models and combinations to deal with the detection of hypernyms from the extracted terminology.
Extracted terminology is evaluated using a manual evaluation carried out by 6 experts in the domain, and the hypernym identification method is evaluated with different datasets. The global approach provides relevant and promising results.
\end{abstract}

\section{Introduction}

The current era is increasingly influenced by the prominence of smart data and mobile applications. The work presented in this paper has been carried out in one industrial project (\VOCA) aiming at automating the production of structured data from human-machine dialogues. Specifically, the targeted application drives dialogues with people working in a construction area for populating a database reporting key data extracted from those dialogues. This application requires complex processing for both transcribing speeches and driving dialogues. The first process ensures good speech recognition in noisy environments. The second processing step is necessary to ensure the database contains accurate and complete data; indeed, people tend to apply a broad (colloquial) vocabulary and the transcribed words must be used to point to the correct information. Additionally, if some data populate the database, additional data may be required for completeness, thus the dialogue should enable getting those additional data (e.g. if the word “room” is recognized and used to populate the database, the location of the room must also be obtained; this can be done by driving the dialogue).

The application provides people with a “hands-free” device, enabling complete, quick, and standardized reporting. The first usage of this application will be oriented to reporting failures and problems in construction site. 

The two processing steps mentioned above require on the one side a “language model” (for transcribing the sentences) and on the other side a “knowledge model” for driving the dialogue and correctly understanding the meaning of the word.
The knowledge model is mainly an ontology of the domain (in this case, the construction domain) providing standardized concepts and their relationships. As well known, building such knowledge models takes time and is costly; one of the earlier questions raised by our industrial partners has been about “how to build, as automatically as possible, such a knowledge model”. This question is closely related to the interest in quickly adapting the application to other domains (than the construction one) for reaching new markets. We developed a complete methodology and system for partially answering the question, focusing on how to extract relevant terminology from a collection of technical specifications.  
The rest of the paper is organized as follows. Section \ref{sec:context} presents the context of the project. Related works are reviewed in Section~\ref{sec:related}. Section~\ref{sec:donnees} presents collected resources and some statistics about them.
Section \ref{sec:MethodologyExtraction} describes the methodology developed for extracting relevant terms from collected resources. 
The details about the evaluation are presented in Section \ref{sec:protocoleExtraction} and the obtained results are given in Section~\ref{sec:ResultatsExtraction}.
Section \ref{sec:MethodologyHypernym} provides details about how processing is introduced for facing the problems of detection of hypernyms and describes different strategies tested.
The details about the evaluation are presented in Section \ref{sec:protocole} and the obtained results are given in Section~\ref{sec:ResultatsMethodologyHypernym}.

\section{Industrial context} \label{sec:context}

\begin{figure}
   \begin{center}
     \centering
     \includegraphics[width=9cm]{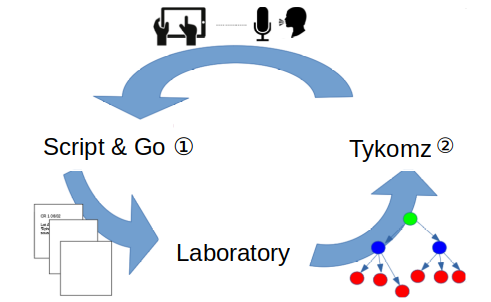}
     \caption{figure describing the context of the project}
     \label{fig:context}
   
   \vspace{-0.5cm}
   \end{center}
 \end{figure}

Figure \ref{fig:context} presents the context of this work in \VOCA~project. Our industrial partner, Script \& Go\footnote{\url{https://scriptandgo.com/fr/}}, developed an application for construction management dedicated to touch devices and wishes to set up an oral dialogue module to facilitate the construction sites data collect.
The second industrial partner, Tykomz, develops a vocal recognition suite based on the Sphynx 4 toolkit\cite[10-19]{Meignier10liumspkdiarization}. 
This toolkit includes agglomerative hierarchical clustering methods using well-known measures such as BIC\footnote{the Bayesian information criterion (BIC)} and CLR\footnote{Cross Likelihood Ratio} and provides elementary tools, such as segment and cluster generators, decoders and model trainers. Fitting those elementary tools together is an easy way of developing a specific diarization system. 
 
\begin{figure*}
   \begin{center}
     \centering
     \includegraphics[width=12cm]{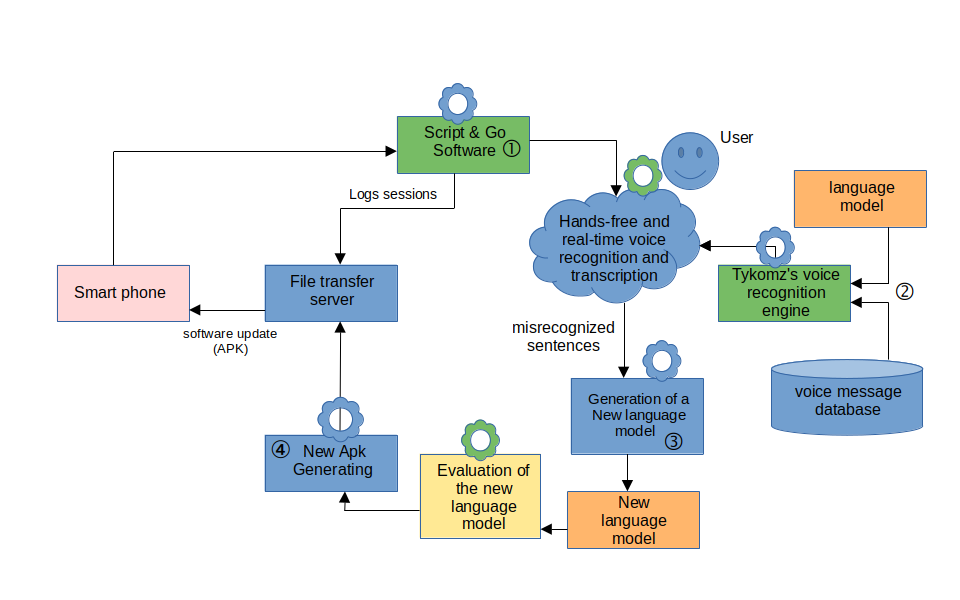}
     \caption{figure describing the final process chain}
     \label{fig:goal}
   
   \vspace{-0.5cm}
   \end{center}
 \end{figure*}

For this application to work, it is necessary to build a model of knowledge, i.e. a model describing the expressions that must be recognized by the program. 
To improve the performance of the system, this knowledge model must be powered by a domain-specific vocabulary.
For example, in the sentence “There is a stain of paint in the kitchen”, the system must understand that it is a stain of paint and that the kitchen is a room. 
Figure \ref{fig:goal} presents the final objective of the VOCAGEN project and describes connections between each partner. 
In the first step, the user downloads the software from Script \& Go (\ding{172}) onto their smartphone or tablet which is connected to the partner's server for requests, logging, and updates. 
Users can use the hands-free module and the software then connects to the voice recognition engine of the second partner to analyze the voice message and transmit a transcription. To perform this transcription, this voice recognition engine uses a language model and a base of test voice messages (\ding{173}). In case of an incorrect voice transcription, the user can manually correct the transcription, and the unrecognized sentences are preserved and analyzed by another process to be able to generate a model allowing these errors to be corrected (\ding{174}). After a step of manual evaluation of the quality of the new language model, a new APK\footnote{Android Package (or APK, for Android Package Kit) is a file format for the Android operating system.} is generated so that the user can download a new version (\ding{175}). 

To our knowledge, there is no ontology or taxonomy specific to the construction industry in French. A version is under development by \cite [100-133]{Pauwels2016} but the ontology is in English and very generic.
We therefore choose to extract useful knowledge from textual data, and then, in a second step, organize it.

\section{Related works} \label{sec:related}

The goal of ontology learning (OL) is to build knowledge models from text.
OL uses NLP knowledge extraction tools to extract terminology and links between them (relationships).
It is concerned with discovering knowledge from various data sources and representing them in an ontological structure. Thus ontology learning comprises a set of techniques to extract the core components of the ontology, i.e. concepts, taxonomic and ad hoc relations, and general axioms. 
For more details, \cite[1-24]{Muhammad2018} provide an overview of existing techniques to accomplish the various subtasks of ontology learning.

\subsection{Taxonomy Extraction} \label{sec:related-term}

Our interest in this part of the work was focused on extracting taxonomic relations. A taxonomic relation occurs between two concepts where a concept is a superordinate of another concept. For example, the concept “Fish” is a superordinate of the concept “Shark” (is-a(Shark, Fish)). We base our approach on the fact that the hypernyms (i.e. relations between terms - a term may comprise several words - in a text representing the fact that a term can be used instead of another term in a sentence, conveying more general or more specific meaning) suggest taxonomic relationships. 
The reference in the field of rule-based systems was developed by \cite[223-254]{Cunningham2002}.  
General Architecture for Text Engineering (GATE) is a Java collection of tools initially developed at the University of Sheffield in 1995.

An alternative is offered by the existing semi-automatic ontology learning system text2onto \cite[227-238]{cimiano2005text2onto}.
More recently, \cite[1–40]{kluegl_toepfer_beck_fette_puppe_2016} developed UIMA, a system that can be positioned as an alternative to GATE. Amongst other things, UIMA makes possible to generate rules from a collection of annotated documents.
Exit, introduced by \cite[946-956]{Roche06} is an iterative approach that incrementally finds the terms.

\cite[49-66]{TERMINAE} with TERMINAE is certainly the oldest statistic approach.
Developed for French and based on lexical frequencies, it requires pre-processing with TermoStat \cite[99-115]{Drouin2003}.
\cite[1-14]{chowdhury2018domain} presents a method for extracting terminology specific to a domain from a corpus of domain-specific text, where no external general domain reference corpus is required. They present an adaptation of the classic \textit{tf-idf} as a ranking measure and use different filters to produce a specific terminology.
More recently, the efficiency of ranking measures like mutual information developed for the statistical approach is discussed in \cite[31-40]{boumaunknownnormalized} and \cite[10-19]{bestgen2017evaluation}. \cite[1-14]{Meyers2018} proposes Termolator, a terminology extraction system using a chunking procedure, and using internet queries for ranking candidate terms. The approach is interesting but the authors emphasize the fact that the runtime for each query is a limiting factor to produce a relevant ranking.

Closer to our work, \cite[55-81]{Gillam05terminologyand} presents an approach combining linguistic patterns and Z-score to extract terminology in the field of nanotechnology. 
\cite[1320-1327]{taxi2016} propose TAXI, which combines statistics and learning approaches with corpus comparison like Termostat. TAXI is a system for building a taxonomy using 2 corpora, a generic and a specific.
It ranks the relevance of candidates by measure (frequency-based), and by learning with SVM.
\cite[496-504]{Lefever2009} and \cite[1-30]{Macken2013TExSISBT} present TexSIS, a bilingual terminology extraction system with a chunk-based alignment method for the generation of candidate terms. After the corpus alignment step, they use an approach combining log-likelihood and Mutual Expectation measures \cite[81–91]{Dias2003AutomaticEO} to rank candidate terms in each language. Similarly, \cite[1-8]{Daille1994StudyAI} and \cite[427-472]{lang2018extracting} present an approach to extract grammatical terminology from linguistic corpora. They compare a series of well-established statistical measures that have been used in similar automatic term extraction tasks and conclude that corpus-comparing methods perform better than metrics that are not based on corpus comparison.
\cite[2-11]{Amjadian2016LocalGlobalVT} and \cite[1-4]{wohlgenannt2016using} present methods with words embedding. With a small data set for the learning phase, they improve the term extraction results in quality of n-gram extracted. 
However, these papers involve labeled data sets for the learning phase, which is the main difference with our proposed approach.
Our approach combine a lexico-syntactical and a statistical approach while using external resources.

\subsection{Detection of hypernyms} \label{sec:related-hypernyms}

Ontology learning refers to the automatic or semi-automatic building of ontology. It is concerned with discovering knowledge from various data sources and representing them in an ontological structure. Thus ontology learning comprises a set of techniques to extract the core components of one ontology i.e. concepts, taxonomic and ad-hoc relations, and general axioms. 
For more details, \cite[1-24]{Muhammad2018} provide an overview of existing techniques to accomplish the various subtasks of ontology learning.

The main approaches found in the literature for extracting hypernyms are pattern-based and distributional, being the latter further divided into unsupervised and supervised.
Pattern-based approaches are heuristic methods that predict hypernym relations between pairs of terms if those terms are related by one pattern matching with a given sentence. These patterns are either defined manually or extracted automatically.
The earliest and most popular handcrafted patterns are introduced by Hearst \cite[1-7]{hearst1992}, thus known as Hearst’s patterns. Pattern approaches can be used for detecting hyperonyms in a text: this is not the case in the work presented in this paper because we extract hypernyms by using a terminology previously extracted and validated. 

Distributional-based approaches are based on the distributional hypothesis, suggesting that words sharing the same linguistic context tend to have a similar meaning \cite[1-6]{harris68}. 
Earlier works to predict hypernymy are unsupervised, usually based on symmetric measures such as cosine similarity \cite[1-15]{Salton1986} and Lin similarity \cite[296-304]{Lin1998}. Various evolutions of earlier works have been proposed such as \cite[1015-1021]{weeds2004}, \cite[425-435]{Shwartz2016}, \cite[1025-1036]{roller2014}, enabling to take into account the inclusion of the context of each term in a pair.

\cite[75-79]{lenci2012} propose to take into account both the inclusion and the non-inclusion of the context of each term in a pair.
Unsupervised methods are simple to implement and apply and there is no need for training data: however, they show low performance and are heavily domain-dependent.

Supervised learning approaches rely on a training dataset to train a model. The
model is then used to predict hypernym relations between terms in a pair. Most of these
    approaches are based on words embedding like Word2vec\cite[3111-3119]{Mikolov2013} or Glove\cite[2249-2259]{Pennington2014}. \cite[970-976]{levy2015} present a study of supervised approaches on standard corpora in the field  and compare state-of-the-art results on labeled datasets.
The various supervised methods differ in the way they represent each candidate
pair of words $(x, y) $: In \cite[2249-2259]{weeds2014} and \cite[1025-1036]{roller2014}, they use the difference between the embedding vector of terms $y$ and the embedding vector of terms $x$ $(~y - ~x )$ as a feature vector to train an SVM classifier. They conclude asymmetric representation performs better and the difference representation yields the best result.
\cite[403-413]{luu2016} proposed an approach to encode hypernym properties by learning terms embedding that not only indicate the information of the hyponym and the hypernyms but also the contexts between them. Consequently, they define triples of hypernym, hyponyms, and the context words between them. They use these triples as training data to build an SVM model by using as feature vectors the concatenation of the embedding vector of hyponyms, the embedding vector of hypernym, and the difference between the two vectors $(~x \oplus ~y \oplus ~y - ~x )$.

According to \cite[579-586]{mirkin2006}, the pattern-based and distributional approaches have
certain complementary properties and they propose a combined approach by learning a supervised model using a set of features obtained by concatenating pattern-based features and distributional-based features. 
In a later work, \cite[425-435]{Shwartz2016} also proposed an approach that combines pattern-based and distributional methods with a supervised model where they learn a neural network model using a concatenation of three feature vectors. A pattern-based feature vector represents the dependency path occurrences of a term pair $(x,y)$ and the other two feature vectors are distributional-based vectors and they represent the embedding vectors of $x$ and $y$.

More recently, pre-trained large-scale language models, such as BERT \cite[4171--4186]{devlin-etal-2019-bert}, RoBERTa\cite[1-13]{liu2019roberta}, and CamemBERT\cite[7203-7219]{martin-etal-2020-camembert} for French,
have proven to be highly effective for many tasks in NLP.
For example, \cite[282-290]{vetter-etal-2022-kamikla} and \cite[260-265]{markchom-etal-2022-uor} propose two similar approaches of fine-tuning the BERT-based models for taxonomic relation classification to tackle one task similar to our at SemEval 2022 \footnote{SemEval is a series of international NLP research workshops focus on semantic analysis \url{https://semeval.github.io/}}.
Recent works on large-scale pre-trained
language models (LLM), such as GPT-3 \cite[1877-1901]{brown2020language}, Llama-2 \cite[510-521]{touvron2023llama} and Claire-Mistral \cite[1-11]{hunter2023claire}, suggest that LLMs also perform well in
various NLP tasks.
Large Language Models (LLMs) feature powerful natural language understanding capabilities. With only a few examples, an LLM can be prompted to perform custom NLP tasks such as text categorization, named entity recognition, information extraction and more.
Close to our task, \cite[5858-5867]{ma-etal-2023-kepl} propose a method for Chinese hypernym relation detection and use the concept of prompt learning to incorporate prior knowledge in the form of patterns into the model.

Although supervised methods overall show better performance than unsupervised ones, they are heavily dependent on the quality and size of the dataset. 
For our purpose, we decided to use supervised methods because they are currently easy to implement, language-independent, transferable to other domains, and compatible with our overall approach where only terminology (not complete texts) is used for detecting hypernyms. By varying approaches, therefore we analyzed which of the state-of-the-art methods performs better in our case.

\section{Terminology Extraction}

\subsection{Resources and statistics}
\label{sec:donnees}

The first experiments were carried out using technical reports\footnote{Site meeting report} collected from three customers from our industrial partners who will be called NC\footnote{For Non-Compliance} collection thereafter. Each document contains all the non-compliance that was found on one work site and describes solutions to resolve it. However, the heterogeneity of the formats, as well as the artificial repetition of the information between two reports found on the same construction site, made the term extraction quite difficult. 
A detailed analysis of these reports reveals a rich vocabulary. However, the presence of numerous misspellings, typing shortcuts, a highly telegraphic style characterized by infinitive verbs, minimal punctuation, and a scarcity of determiners significantly hinders the exploitation of this vocabulary.
As a consequence, we used a collection of technical specifications called CCTP\footnote{The technical specifications book (CCTP in French) is a contractual document that gathers the technical clauses of a public contract in the field of construction.}. {CCTPs are available online on public sector websites\footnote{For example, https://www.marches-publics.gouv.fr/ or http://marchespublics.normandie.fr/.}. Several thousand documents were collected by our industrial partner using an automatic web-collecting process.
Figure \ref{tab:statistics} presents some key descriptive statistics of these collections. 
\begin{figure}[ht] 
\begin{center}
\begin{tabular}{|l|r|r|} 
\hline
Collection&\multicolumn{1}{|c|}{NC}&\multicolumn{1}{|c|}{CCTP}\\ \hline
Total number of documents  &58~402& 3665	\\ \hline 
\multicolumn{3}{c}{\emph{Without pre-processing}} \\\hline
Total number of words &130~309&230~962~734 \\ \hline
Total number of different words&93~000 &20~6264 \\ \hline
Average words/document &125.3&63~018.48 \\
\hline
\end{tabular}

\caption{statistics of the collection.}
\label{tab:statistics}
\end{center}
\end{figure}

\subsection{Methodology of Extraction of Terminology}
\label{sec:MethodologyExtraction}

\subsubsection{System Overview}
Figure \ref{fig:overview} presents an overview of the system designed and implemented: steps are explained in further sections. In Step 1 pre-processing of raw information extracted from CCTP collection takes place; this is required for normalizing the entire set of documents.
In Step 2, n-grams are extracted (by using measures). 
1,2,3 grams are extracted. 
In Step 3, n-grams are filtered by using linguistic patterns and Internet queries. Finally, in Step 4 a ranking is applied to the filtered n-grams.

\begin{figure*}
   \begin{center}
     \centering
     \includegraphics[width=12cm]{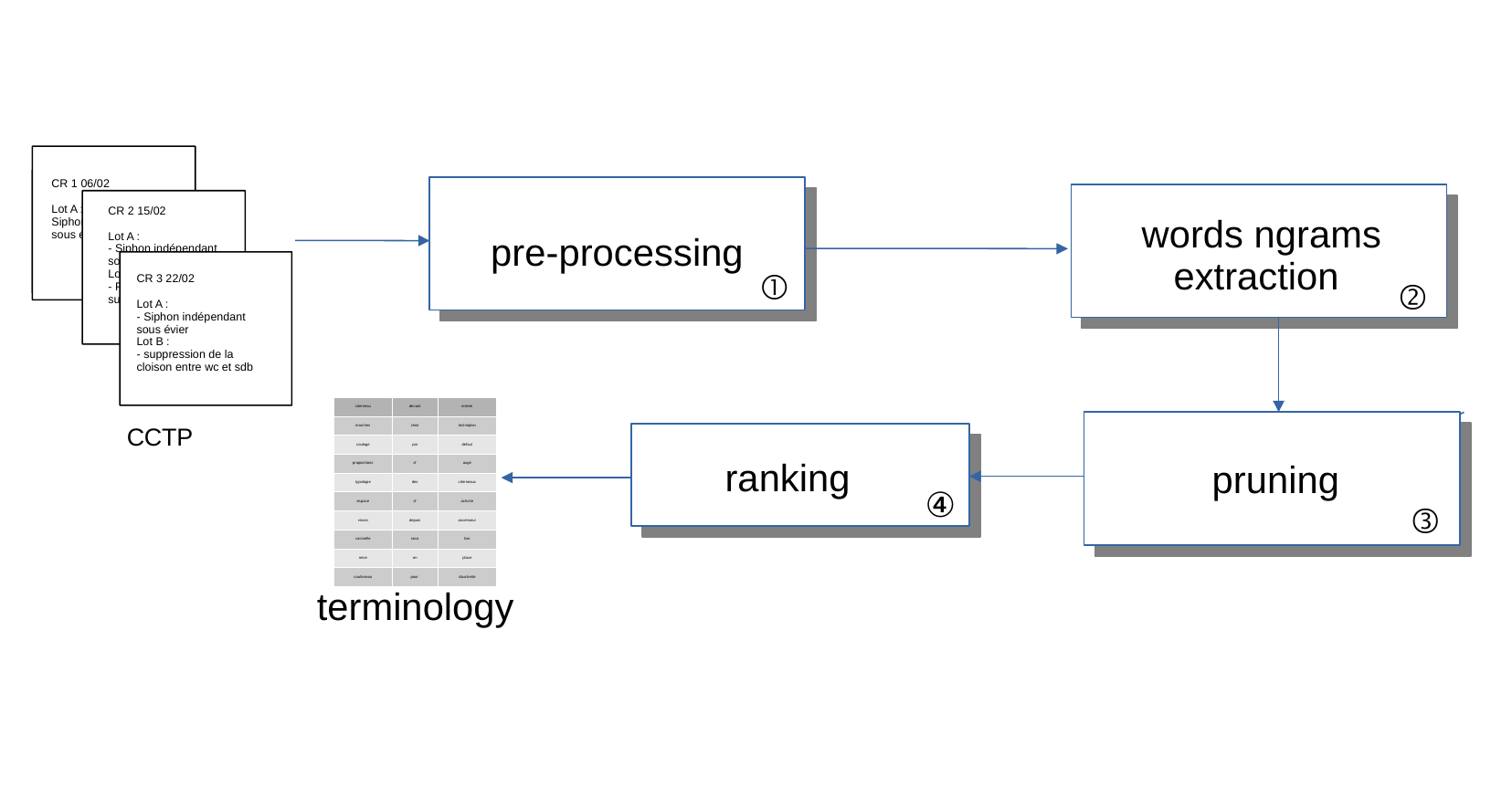}
     \vspace{-0.5cm}
     \caption{System overview}
     \label{fig:overview}
   
   \vspace{-0.5cm}
   \end{center}
 \end{figure*}
 
\subsubsection{Normalization, pre-processing and word n-grams extraction}
\label{sec:processing}

In Step 1, a text normalization is performed to improve the quality of the process. We remove special characters such as ``/'' or ``()''.
Different pretreatments are done to reduce noise in the model: we remove numbers (numeric and/or textual) and special symbols. ``.'' are tagged with a special character to not create artificial n-grams. Specific words (including named entities) like company names, dates, etc. are normalized and will be removed in the next module. We do not include a stop list to keep n-grams with prepositions, for the purpose described in the remainder.
Then, we tokenize the entire collection before using TreeTagger \cite{schmid1994probabilistic} to get the part-of-speech tags and lemmas of each word.
After this step we transform all vocabulary from the CCTP collection into 1-grams, 2-grams and 3-grams. Special characters or normalized words resulting from the previous processing are discarded. N-grams with a very low frequency (2) are also discarded. 

\subsubsection{Linguistic patterns module}

We use grammatical labels generated in the previous step (section \ref{sec:processing}) and linguistic patterns to retrieve collocations such as NOUN-NOUN and NOUN-PREP-NOUN. These patterns are frequently found in the literature\cite[946-956]{Roche06} to capture specific words in French like “carte de cr\'{e}dit”(\textit{credit card}) and discard 3-gram like  “cr\'{e}diter sa carte” (\textit{credit his card}) with the pattern VERB-PREP-NOUN.
Among frequent patterns found in literature, those patterns have been selected according to the statistics obtained from a knowledge model of another field (agriculture), given by one of our industrial partners. Figure \ref{stats:modele} presents the main patterns we selected using this knowledge model. 
\begin{figure}[!ht]
 \centering
 \begin{tabular}{lcc} 
 \hline
&Number&Percentage\\ \hline
 1-grams&1360&65.24\%\\\hline
 NOUN&1037&76.25\%\\
 VERB&194&14.26\%\\
 ADJ&120&8.82\%\\\hline
  2-grams&390&19.57\%\\\hline
 NOUN-NOUN&346&88.72\%\\
 ADJ-NOUN&11&2.82\%\\
 PREP-NOUN&7&1,79\%\\
 VERB-NOUN&5&1,28\%\\\hline
 3-grams&188&9.43\%\\\hline
 NOUN-NOUN-NOUN&150&79.79\%\\
 PREP-NOUN-NOUN&15&7.98\%\\
 NOUN-PREP-NOUN&6&3,19\%\\
 VERB-NOUN-NOUN&6&3,19\%\\
 \end{tabular}
 \caption{Distribution of linguistic patterns according to the knowledge model.}
 \label{stats:modele}
 \end{figure}
The sum of percentages may not reach 100\% as infrequent patterns are excluded.
We observed that the noun-based patterns are the most frequent patterns, whatever the size of the n-gram.
The other selected patterns also contain nouns, but they are n-grams with verbs, adjectives or prepositions. Therefore, we have configured our system to keep only the n-grams corresponding to these patterns.

\subsubsection{Pruning step}

This step uses the Internet to prune n-grams for which no information is returned after querying Bing\footnote{\url{https://www.bing.com/}} search engine.
We count the number of links in the result pages that contain the n-gram exactly.
We save the number of exact matches between the n-gram and the title and snippet of each result.
We keep only the n-grams whose number of matches exceeds a defined threshold. 
We varied this threshold between 1 and 50 and results presented in Section  \ref{sec:ResultatsExtraction} have been obtained with a threshold empirically set to 10. 

\subsubsection{Ranking step}
\label{sec:ranking}

We tested several measures as provided in \cite[946-956]{Roche06} and \cite[1-30]{Macken2013TExSISBT} like mutual information in order to rank selected n-grams by quality but the results were disappointing. We finally use classical \emph{Z score} \cite[589-609]{Altman1968} with twenty years of the French newspaper Le Monde\footnote{\url{http://www.islrn.org/resources/421-401-527-366-2/}} as a generic collection. This metric considers word frequencies weighted over two different corpora, in order to assign high values to words having much higher or lower frequencies than expected in a generic collection.  We defined it as follows :

\begin{equation}
   \label{eq:Zscore:p_1}	
   p_1 = a_0 / b_0 
\end{equation}	 
\begin{equation}
   \label{eq:Zscore:p_2}	
   p_2 = a_1 / b_1	 		
\end{equation}	 
\begin{equation}
   \label{eq:Zscore:p}	
    p = (a_0 + a_1) / (b_0 + b_1)	 		
\end{equation}	 
\begin{equation}
   \label{eq:Zscore}	
    Z_Score=\frac{p_1 - p_2}{ \sqrt{(p*(1 - p)*(\frac{1}{b_0}+ \frac{1}{b_1})}}
\end{equation}	 

\noindent Where $a$ is the lexical unit considered (1-gram, 2-gram or 3-gram), $a_0$,$a_1$  the frequency of $a$ in the CCTP collection, $b_0$ the total size in words of CCTP collection, $b_1$ the frequency of $a$ in the collection Le Monde.

\subsection{Experiments and results}
\label{sec:xpresultats}

\subsubsection{Experimental protocol}
\label{sec:protocoleExtraction}

To assess their quality, we manually evaluated all 3-grams the system retained. Six construction specialists participated, each evaluating approximately one third of the data. 5144 3-grams were evaluated with this method and each n-gram was evaluated by 2 different specialists\footnote{With a inter-annotator agreement of 0.83}. For each n-gram, the specialist can choose between three possibilities:

\begin{enumerate}
\item the 3-gram is irrelevant ;
\item the 3-gram is relevant but does not belong to the domain\footnote{For example, credit card (carte de crédit in french) is a relevant n-gram but does not belong to the domain of construction} ;
\item the 3-gram is relevant and belongs to the domain.
\end{enumerate}

The evaluation was done in two steps and we use Kappa measure\footnote{We use general formula as follows: $\kappa=\frac{A_0-A_e}{1-A_e}$ where $A_0=$ observed agreement and $A_e=$ expected (chance) agreement. } \cite[1-10]{cohen1960} and inter-annotator agreement at the end of the first step to show the difficulty of the task. At the end of the first step, we obtained a Kappa score of 0.62 and a global inter-annotator agreement of 0.74, which is quite good as explained in \cite[1-7]{McHugh2012InterraterRT}. The difficulty of the task was to distinguish the domain-specific vocabulary from the generic vocabulary used in the field of construction.
Each disagreement was re-evaluated in the second step by a pair of experts. Figure~\ref{eval:vocagen3G} shows the final results of the evaluation.

\subsubsection{Results}
\label{sec:ResultatsExtraction}

In this section, we present the results obtained during the manual evaluation of the 3-grams retained by the system. 
We only compute the accuracy and the error rate, because we are not able to compute the recall for this collection\footnote{Indeed, we do not know every relevant terms existing in the corpus, so we cannot estimate the recall for the collection of terms we automatically extract.}.
We have merged the assessments of each expert using two different evaluation rules:

\begin{itemize}
\item a strict evaluation where an n-gram is considered correct if both experts have rated it relevant and in the domain.
\item a flexible evaluation where an n-gram is considered correct if both experts consider it relevant and at least one of the experts considers it the domain.
\end{itemize}

\begin{figure}[!ht]
 \centering
 \begin{tabular}{lccc} 
 \hline
&strict evaluation&flexible evaluation\\ \hline
accuracy  &0.77&0.91\\
error rate  &0.23&0.09\\
 \end{tabular}
 \caption{Results of manual evaluation on the 3-grams.}
 \label{eval:vocagen3G}
 \end{figure}
 
The strict evaluation shows good quality results (0.77). Analysis of the results shows that the main error is related to “incomplete n-grams”. For example, the 3-gram ``personne \`{a} mobilit\'{e}'' (person with mobility) is not relevant while the 4-gram ``personne \`{a} mobilit\'{e} r\'{e}duite'' (person with reduced mobility) can belong to the field of construction. 
Some errors can also be traced back to the CCTP documents. For example, ``engin de guerre'' (war machine) is a term that does not belong to the field but a law relating to the presence of a war machine on the building sites is reported in every CCTP.
 The flexible evaluation shows very good results (0.91) and the difficulty of assessing the class of some terms such as ``absence de remise'' which has 2 distinct meanings in French (no outhouse and no discount). The first meaning is relevant in the field of construction but not the second.
This part of the work was previously published in \cite[22-26]{Kessler2019}. 
After the terminology extraction step, we worked on a model to detect hypernym relationships, because this task was far too time-consuming to do manually with all the extracted terms.

\section{Detection of Hypernym Relationships}
\label{sec:MethodologyHypernym}

This section focuses on relation extraction from the terminology previously obtained. Two methodologies are studied, the first use classical words embedding vectors as features, and the second one relies on end-to-end approaches. To validate the methodology, two other datasets are experimented with.

\subsection{Methodology}

\subsubsection{Words embedding-based module}

\paragraph{Overview}
\begin{figure}
   \begin{center}
     \centering
     
     \includegraphics[width=9.5cm]{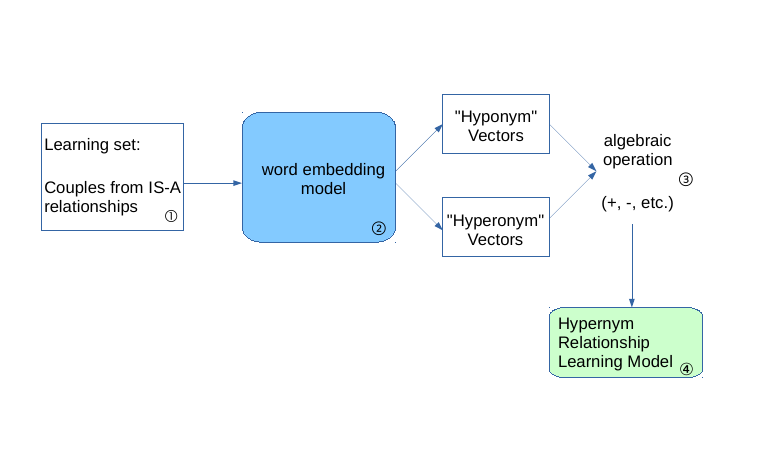}
     \caption{Architecture of the hypernym detection module}
     \label{fig:overviewRelationShip}
   
   \vspace{-0.5cm}
   \end{center}
 \end{figure}

Figure \ref{fig:overviewRelationShip} presents the architecture of the Words embedding-based module. In the first step, we use a training dataset (\ding{172}) composed of pairs of terms that are known to be in a hypernym relationship (or hyponym relationship depending on the reading order). For instance, the pair “kitchen, room” represents a hyponym relationship because a kitchen is a room. 
We combine this training dataset with a words embedding model (\ding{173}) providing a vector for each term of each pair in the training dataset. A vector algebraic operation (\ding{174}) is then applied between each word of the pair to create a hypernym relationship learning model (\ding{175}).
Such kind of model is fit using classical machine learning algorithms like Random Forests or Multilayer Perceptron. Finally, the obtained model is used to detect hypernym relationships in the extracted terminology.

\paragraph{Embedding models component}

Words embedding is a representation technique to represent any word in the low-dimensional space:  words having similar representations are likely to be semantically similar. 
Each word found in an input dataset is projected to a vector model to obtain a semantic representation of the whole input dataset.  We experimented with three words embedding models as follows.  
The first model is Word2vec \cite[3111-3119]{Mikolov2013}, an unsupervised and predictive neural words embedding technique to learn the word representation in low-dimensional space. We specifically use the skip-gram model, in which a word is used to predict the context using a neural network close to an autoencoder.
Second is Glove, for Global vector for Word Representation \cite[1532--1543]{Pennington2014}. A co-occurrence word matrix is created from a text dataset for the training and is reduced in low-dimensional space which explains the variance of high-dimensional data and provides a word vector for each word.
The last model is fastText \cite[135-146]{bojanowski2017enriching}, which is close to Word2vec. Unlike Word2vec which considers each word as a single unit and ignores the morphological structure of the word, fastText overcomes this limitation by considering each word as an n-gram of characters. 

Note that we do not mention in our experimental section the results obtained with Word2vec because they were systematically lower than those obtained with fastText.

\paragraph{Vector compositions and machine learning algorithms}

We tested four compositions for representing $(x, y)$ as a feature vector: \textbf{concat} $(~x \oplus  ~y)$ \cite{baroni2010}, \textbf{diff} $(~y - ~x)$ \cite{weeds2014}, \textbf{sum} $ v_b + v_a $ and \textbf{product} $ v_b * v_a $. 
We evaluated the performance of the following classification algorithms: SVM (Support-Vector Machine), RF (Random Forests), and MLP (Perceptron Multilayer), and two fusions of vote-based algorithms: the fusion so-called Hard (the majority prevails) and the so-called Soft (we sum the probabilities of prediction). However, we have got bad results with fusion and have not included results in the paper. 

\subsubsection{End-to-end based module}

\paragraph{Overall}

Unlike the module presented previously based on classic words embedding, an end-to-end approach can be defined generically as a system that processes the entirety of a task in a manner unified, without manual or decoupled intermediate steps. In effect, the system takes raw data as input and produces the desired result, without requiring human interventions or third-party systems with the possible exception of minimal pre/post-processing. Usually, we use a pre-trained model to implement an end-to-end approach. 

For our task, such an approach amounts to using two words as input to the model (the hyponym and the hypernym) and as the expected output the class label (i.e. in relation of hyperonymy or not). Thus, the model can be fine-tuned with the training dataset. We do not need to use precomputed embedding like Word2vec because the model will learn contextualized embedding weights during the training process.

\paragraph{Used algorithms}

We evaluated the performance of three algorithms which can be used in an end-to-end process. A mask-based algorithm RoBERTa (and this French version CamemBERT), and two causal models (which are also Large Language Models) Llama-2 and Claire-Mistral. 

RoBERTa (Robustly Optimized BERT Pretraining Approach) is a pre-trained Transformers-type language model, developed by Facebook in 2019. It is an optimized and improved version of the BERT model, with pre-training on many more data.

LLaMA-2 (Longform Language Model with Attention) is a Large Transformers-type Language Model, recently developed by Facebook in 2023. We use here the 7B model.

CLAIRE/Mistral is a Large Language Model developed by French researchers in 2023 adapted from Mistral. Built on the Transformers architecture, it was pre-trained specifically on diarized French conversations. 

Experimental protocol and results are presented in the next section. 

\subsection{Experiments and results}
\label{sec:xpresultatsEmbedding}

\subsubsection{Datasets}
\label{sec:dataset}

We rely on three datasets of semantic relations, which are all used in various
state-of-the-art approaches, for hypernym’s evaluation. 
The first dataset is BLESS \cite[23-32]{Baroni2012EntailmentAT}, which contains 1,337 hyponym-hypernym pairs. It is designed to evaluate distributional semantic models. It contains 200 distinct concepts. Concepts are named with single-word nouns in the singular form. Each concept has
a set of related words. A concept and related word are labeled by one of the following
five relations: co-hyponym, hypernym, meronym, attribute, and events. For instance, the dataset contains “cat-hyper-animal” where “cat” is the
concept, “animal” is the related word, and “hyper” is the label. 
The second data dataset is EVALution, introduced by \cite[64-69]{santus2015}. It is also designed for the purpose of training and evaluating distributional semantic models. It consists of 7,500 couples representing the following semantic relationships: hypernymy, synonymy, antonymy, and meronymy. Couples were extracted from ConceptNet 5.0 and WordNet 4.0 \cite[211-226]{Liu2004}.
The latest dataset, VOCAGEN, written in French, contains 4,143 couples, specific to the construction domains, extracted from the \Tykomz~partner knowledge model.

For each dataset, we define, if not available, \textit{positive couples} (the ones containing hyponymy relationships), and \textit{negative couples} (the ones not containing hyponymy relationships). BLESS dataset already contains negative couples. For the two other datasets, we use the following process to define negative couples. We first randomly select a list of hypernyms noted $H$ from the positive couples. Then, we add to each $H$ element a fixed $n$ number of words that we did not get with the hypernym before, in order to be sure to produce a negative couple. In the end, we fix the value of $n$ in order to have a balanced dataset (the same number of positive and negative couples).

\subsubsection{Words embedding used models}

We have used generic models for English and French. We originally built a specific model based on technical documents collected by our industrial partner using an automatic web-collecting process and described in section \ref{sec:donnees}, but the results were very disappointing and were not included in the paper. 

Pre-trained embedding models for English datasets BLESS and EVALution are the following.
\begin{itemize}

    \item Two GloVe models are used:
    \textbf{glv\_core}\footnote{\url{https://spacy.io/models/en##en_core_web_lg}}
     and \textbf{glv\_vect}\footnote{\url{https://spacy.io/models/en-starters##en_vectors_web_lg}}. These models are included in the SpaCy python module.  glv\_core is the embedding model from SpaCy called “en\_core\_web\_lg”. This model was trained with GloVe from the Common Crawl dataset (\url{https://commoncrawl.org/}). glv\_vect is also trained with GloVe from blogs, news, and comments.
     \item The \textbf{fastText} vectors \cite[1-4]{mikolov2018advances} contain 2 million word vectors trained on Common Crawl and are available on the official web page of the authors\footnote{\url{https://fasttext.cc/docs/en/crawl-vectors.html}}.
\end{itemize}

For the VOCAGEN dataset, we use the French version of these pre-trained embeddings.

\subsubsection{Machine learning used models}

We use the scikit-learn\footnote{\url{https://scikit-learn.org/}} implementation for classical machine learning algorithms SVM (Support-Vector Machine) RF (Random Forests), and MLP (Multilayer Perceptron). We also experiment TPOT\footnote{\url{https://epistasislab.github.io/tpot/}} which is an auto-classification models tool that combines different statistical analyses and machine learning algorithms to get the best possible model.
The used parameters for these different algorithms are the following:
\begin{itemize}
    \item MLP: We use two layers of size 100 with a ReLU activation, an Adam solver, a batch size of 32 with a maximum iteration number of 100, a random state of 42, and an early stopping with a tolerance of 0.0001.
    \item RF: We use 1,000 estimators, a max depth of 1,000, and a random state of 42.
    \item SVM: meta parameters of SVM are fixed using a cross-validated grid search on the training dataset. Tested parameters values are C: .1, 1, 10, 100, 1,000; gamma: (scale, auto); and kernel: rbf, sigmoid.
    \item TPOT: We used pre-implemented scikit-learn algorithms with TPOT with the following parameters: number of generations: 10, a population size of 100 and a random state of 42.
\end{itemize}

In the case of other models, we use  the Hugging Face\footnote{\url{https://huggingface.co/}} implementation and the \texttt{AutoModelForSequenceClassification} class to add an appropriate output head to the model.
For CamemBERT/RoBERTa, we use the base and large versions of the models. RoBERTa is used for English datasets and CamemBERT for the French dataset. The following parameters are used: a learning rate of 5e-5, a batch size of 64, and a maximum of 100 epochs (using an early stopping patience of 5). Other parameters are the default ones. With the large version of these models, we change the learning rate to 1e-6.
We experiment with the \texttt{Llama-2-7b-hf} and the \texttt{Claire-Mistral-7B-0.1} models. The following parameters are used: a 4-bit quantization is made by using the \texttt{bitsandbytes} module, adding an adapter for fine-tuning using Lora with \texttt{r=16}, \texttt{lora\_alpha=32} and all target modules. We use a batch size of 16 and a maximum of 100 epochs (using an early stopping patience of 5). The learning rate is fixed to 2e-5.

\subsubsection{Experimental protocol}
\label{sec:protocole}

The experimental protocol is as follows. The objective is to check the quality of the prediction model for each selected dataset. Only the positive class is evaluated, that is to say that we only search if the couples judged to be hypernym relations are so.
Each of the experiments is evaluated using the well-known measures of Precision, Recall, and F-measure, averaged over all classes (with $ \ beta = 1 $ in order not to privilege precision or recall \cite[345-359]{Gaussier}). 
The results were obtained using 5-fold cross-validation on each dataset\footnote{
 We use this distribution of data : BLESS train 1000, dev 334, test 334, EVALution : train 2256, dev 752, test 752}. The classifier performance is shown in the table below. The table includes the average F1-score across the 5 folds (mean) and the standard deviation ($\pm$SD) for each measure.
Results are finally compared to a baseline model. 

\subsubsection{Results}
\label{sec:ResultatsMethodologyHypernym}

\paragraph{State-of-the-art datasets results}
\label{results:BLESS}

Figures \ref{eval:BLESS2} and \ref{eval:EVALution2} present a summary of the results obtained with each classification algorithm and each model on both BLESS and EVALution datasets.

\begin{figure}[h]
\small
        \centering
\begin{tabular}{|c|c|c|c|c|} 
 \cline{1-5}
Algo&Emb/Size&$\mu$ F1&$\mu$ R&$\mu$ P\\ \hline
Llama-2&7B&\textbf{97.81}$\pm$1.14&98.80$\pm$0.93&96.86$\pm$1.84\\ \hline
RF&FastText&97.55$\pm$0.90&95.69$\pm$1.98&99.51$\pm$0.46\\ \hline
SVM&Glove&97.54$\pm$0.42&97.48$\pm$0.70&97.62$\pm$1.10\\ \hline
Mistral&7B&97.31$\pm$0.64&97,60$\pm$0.66&97.04$\pm$1.21\\ 
\hline
TPOT&FastText&97.19$\pm$0.55&95.45$\pm$1.72&95.45$\pm$0.90\\ \hline
RF&Glove&97.10$\pm$0.91&94.73$\pm$1.98&99.63$\pm$0.49\\ \hline
C.-M.&7B&97.08$\pm$1.09&97.72$\pm$1.22&97.72$\pm$1.12\\ 
\hline
TPOT&Glove&97.05$\pm$1.13&96.05$\pm$0.96&98.10$\pm$2.30\\ \hline
RoB&large&96.97$\pm$1.05&97.96$\pm$0.98&96.01$\pm$1.24\\
\hline
MLP&Glove&96.68$\pm$0.99&97.48$\pm$0.70&95.88$\pm$1.31\\ \hline
RoB&Base&96.56$\pm$0.63&96.04$\pm$1.23&97.10$\pm$0.68\\ \hline
MLP&FastText&96.04$\pm$1.09&97.24$\pm$1.04&94.87$\pm$1.41\\ \hline
SVM&FastText&95.99$\pm$0.91&97.36$\pm$0.48&94.66$\pm$1.48\\ \hline

\end{tabular}
\caption{Precision, recall and f1-score obtained on BLESS dataset.}
\normalsize

\label{eval:BLESS2}
\end{figure}

We observe that all the algorithms achieve good results for each of the two datasets. 
However, on Bless dataset, all the algorithms obtain very good results (F-score between 95.99 and 97.81) while performances are more unequal on the EVALution dataset (F-score between 65.18 and 81.84).

 \begin{figure}[h]
        \centering
        \small
\begin{tabular}{|c|c|c|c|c|} 
 \cline{1-5}
Algo&Emb/Size&$\mu$ F1&$\mu$ R&$\mu$ P\\ \hline
Claire-M&7B&\textbf{81.85}$\pm$3.25&79.47$\pm$9.07&85.77$\pm$5.97\\ 
\hline
Mistral&7B&79.80$\pm$3.34&77.87$\pm$9.54&83.39$\pm$5.87\\ 
\hline
Llama-2&7B&80.62$\pm$3.91&79.63$\pm$7.87&82.22$\pm$2.53\\ \hline
RoB&large&81.15$\pm$2.5&84.73$\pm$2.69&78,06$\pm$4.52\\
 \hline
RoB&Base&80.28$\pm$1.28&83.46$\pm$5.27&77.73$\pm$3.43\\ \hline
SVM&Glove&73.36$\pm$0.48&72.77$\pm$0.85&74.00$\pm$1.63\\ \hline
TPOT&FastText&72.70$\pm$0.91&72.66$\pm$1.18&72.73$\pm$0.71\\ \hline
TPOT&Glove&72.60$\pm$1.84&72.23$\pm$3.39&73.07$\pm$1.14\\ \hline
MLP&FastText&71.03$\pm$1.78&70.43$\pm$2.84&71.70$\pm$1.51\\ \hline
MLP&Glove&69.32$\pm$2.09&69.47$\pm$3.82&69,31$\pm$1.82\\ \hline
SVM&FastText&68.78$\pm$1.42&68.03$\pm$1.74&69,56$\pm$1.45\\ \hline
RF&Glove&65.70$\pm$1.30&64.04$\pm$1.80&67.47$\pm$1.47\\ \hline
RF&FastText&65.19$\pm$1.50&63.88$\pm$1.49&66.56$\pm$1.81\\
\hline

\hline

\end{tabular}
\normalsize
\caption{Precision, recall and f1-score on EVALution dataset.}
\label{eval:EVALution2}

\end{figure}
This performance difference likely stems from the EVALution corpus encompassing diverse relationship types, whereas we solely focused on 'is-a' relationship. This limited our training data for experimentation and fine-tuning.
On both datasets, the best results were obtained by LLMs: Llama-2 on BLESS and Claire-Mistral on EVALution, using 7B embedding models for both. 
We also tested Mistral for text generation. However, the results were quite similar to those achieved by Claire-Mistral and Mistral in classification tasks (on BLESS 97.23 versus  97.08 and 97.31 respectively).
Even though variations remain limited, we observe an influence of embedding models on each algorithm.

While algorithms using pre-trained vector representations(RF, SVM, MLP) achieved excellent results on the Bless dataset, their performance dropped significantly on the second dataset.
As previously discussed, the EVALution dataset may present greater processing challenges due to its inherent complexity.
TPOT, RoBERTa achieve good results on the two datasets with slight variations depending on the chosen embedding.

 \begin{figure}[h]
        \centering
        \small
\begin{tabular}{|c|c|c|c|c|} 
 \cline{1-5}
Algo&Emb/Size&$\mu$ F1&$\mu$ R&$\mu$ P\\ \hline
CBert&Large&\textbf{91.61}$\pm$0.74&93.63$\pm$1.11&89.69$\pm$1.03\\ \hline
CBert&Base&90.97$\pm$0.96&93.35$\pm$2.18&88,75$\pm$1.14\\ \hline
C.-M.&7B&88.72$\pm$1.41&91.32$\pm$4.20&86.50$\pm$2.89\\ \hline
Llama-2&7B&88.17$\pm$1.54&89.35$\pm$2.48&87.28$\pm$4.65\\ \hline
TPOT&FastText&87.17$\pm$0.61&88.57$\pm$1.96&85.86$\pm$1.38\\ \hline
SVM&FastText&84.91$\pm$0.82&84.80$\pm$1.4&85,03$\pm$0.38\\ \hline
MLP&FastText&84.34$\pm$1.21&84.73$\pm$2.80&84.17$\pm$3.59\\ \hline
RF&FastText&80.53$\pm$1.57&76.87$\pm$2.34&84.60$\pm$0.91\\ \hline
TPOT&Glove&79.92$\pm$0.51&81.42$\pm$1.46&78.51$\pm$1.01\\ \hline
SVM&Glove&77.90$\pm$0.96&78.18$\pm$2.19&77.67$\pm$0.76\\ \hline
MLP&Glove&77.04$\pm$0.89&79.24$\pm$2.14&74.99$\pm$0.44\\ \hline
RF&Glove&74.25$\pm$1.88&70.84$\pm$2.56&78.03$\pm$1.12\\ \hline

\end{tabular}
\normalsize
\caption{Precision, rappel and f1-score obtained on \VOCA~ dataset.}
\label{eval:vocagen}
\end{figure}

\paragraph{\VOCA~dataset results}
\label{results:VOCA}

Figures \ref{eval:vocagen} present results obtained with each classification algorithm and each model on \VOCA~dataset.
The best results are obtained with Camembert (CBERT in table), slightly better for large embedding than for base embedding (respectively 91.61 and 90.97) with a very small standard deviation.
Llama-2 and Claire-Mistral confirm the good results obtained on the literature datasets with very good scores as well.
We observe that the scores of linear algorithms are slightly lower, as for the evaluation dataset, which confirms the difficulty of our task. There are slight differences depending on the embedding model used but the best performance is obtained with fastText. 

A preliminary analysis was carried out on the best results obtained. These results showed different types of possible errors.
For example, the terms “coiffeuse” (vanity table) and “meuble” (table) are an interesting error. In French, “coiffeuse” can refer to both a profession (hairdresser) and a specific type of furniture (vanity table). This ambiguity can explain why the model makes mistakes.
More generally, a small portion of the relationships contained in the dataset are not IS-A type relationships, but rather Part-of type relationships or others, for example “Europe” and “Eurasia” or “moldy” and “mushroom”.
A cleaning step is underway to remove these relationships.
Most classification errors stem from complex technical terms or terms that have different meanings in the context of construction. For example, “poutre longrine” translates to “longrine beam”, a specific type of beam used in construction. Similarly, “tuyau dauphin” is a French technical term for a type of flexible hose used for suction and discharge of water, although it literally translates to “dolphin hose”.
The last category of errors involves brand mentions within relationship terms. For example, “Bic pen” or “Merlin axe”, here Merlin is a brand of an axe. In our context, the model needs to understand sentences like “put away the Merlins” and identify that the user is referring to axes.

\section{Conclusion and future work}
\label{sec:conclusion}

The paper reports our experiments and results for building a precise and large terminology for the construction domain. Collecting terminology is indeed the first step towards a complete knowledge model containing both concepts and relationships. 
During our work we were faced to several problems: finding resources and selecting them for building an appropriate corpus, thinking and developing pre-processing for cleaning those resources, experimenting distinct measures for n-grams and selecting the most appropriate, improving results by adding linguistic patterns and Internet queries. The current results are quite promising according to the evaluation of the extracted terminology carried out by 6 experts in the field. 
However, as manual evaluation of the produced terminology proved to be time-consuming and laborious, we developed in a second time a model to automatically or semi-automatically system to validate the terminology using our second partner's knowledge model. Our goal was to propose a module for the detection of hypernyms performing and different combinations were tested for different corpora/languages.
We explore two approaches for representing text data, the first uses classical words embedding vectors as features, and the second one relies on end-to-end approaches.
Models fine-tuned with CamemBERT achieved a very promising F-score of 91.6\%.
While construction terminology may appear specialized and technical, deep learning approach with LLM and BERT model trained on a massive dataset seems to be sufficient for this task.
From our French hypernymy relation model trained on VOCAGEN dataset and validated on the two other datasets, the next step will be to organize the extracted terminology to build a significantly richer taxonomy.
This work validates the possibility of changing domains (a key objective for one of the partners) and maintaining good performance.
We propose first improving results by merging the outputs of all algorithms, assuming they make errors on different data. A more in-depth analysis of the errors is necessary to validate this hypothesis.
 Another interesting perspective for this work would be to explore using the knowledge model from our partner as input to enhance the model's prediction capabilities.


\section*{Acknowledgement} The authors would like to thank Mickaël Kessler for his help for reviewing and entering the results.

\bibliography{Terminology2024}

\begin{thebibliography}{}

\bibitem[\protect\citeauthoryear{Altman}{Altman}{1968}]{Altman1968}
Altman, E. (1968).
\newblock Financial ratios, discriminant analysis and the prediction of
  corporate bankruptcy.
\newblock In {\em The Journal of Finance, 23(4). doi:10.2307/2978933}, pp.\
  589--609.

\bibitem[\protect\citeauthoryear{Amjadian, Inkpen, Paribakht, and
  Faez}{Amjadian et~al.}{2016}]{Amjadian2016LocalGlobalVT}
Amjadian, E., D.~Inkpen, T.~S. Paribakht, and F.~Faez (2016).
\newblock Local-global vectors to improve unigram terminology extraction.
\newblock In {\em Proceedings of the 5th International Workshop on
  Computational Terminology}, pp.\  2--11.

\bibitem[\protect\citeauthoryear{Asim, Wasim, Khan, Mahmood, and Abbasi}{Asim
  et~al.}{2018}]{Muhammad2018}
Asim, M.~N., M.~Wasim, M.~U.~G. Khan, W.~Mahmood, and H.~M. Abbasi (2018, 10).
\newblock {A survey of ontology learning techniques and applications}.
\newblock {\em Database\/}~{\em 2018}, bay101.
\newblock bay101.

\bibitem[\protect\citeauthoryear{Baroni, Bernardi, Do, and chieh Shan}{Baroni
  et~al.}{2012}]{Baroni2012EntailmentAT}
Baroni, M., R.~Bernardi, N.-Q. Do, and C.~chieh Shan (2012).
\newblock Entailment above the word level in distributional semantics.
\newblock In {\em EACL}, pp.\  23--32.

\bibitem[\protect\citeauthoryear{Baroni and Lenci}{Baroni and
  Lenci}{2010}]{baroni2010}
Baroni, M. and A.~Lenci (2010).
\newblock Distributional memory: A general framework for corpus-based
  semantics.
\newblock {\em Computational Linguistics\/}~{\em 36\/}(4), 673--721.

\bibitem[\protect\citeauthoryear{Bestgen}{Bestgen}{2017}]{bestgen2017evaluation}
Bestgen, Y. (2017).
\newblock {\'E}valuation de mesures d’association pour les bigrammes et les
  trigrammes au moyen du test exact de fisher (using fisher’s exact test to
  evaluate association measures for bigrams and trigrams).
\newblock In {\em Actes des 24{\`e}me Conf{\'e}rence sur le Traitement
  Automatique des Langues Naturelles. Volume 2-Articles courts}, pp.\  10--18.

\bibitem[\protect\citeauthoryear{Bi{\'e}bow, Szulman, and
  Cl{\'e}ment}{Bi{\'e}bow et~al.}{1999}]{TERMINAE}
Bi{\'e}bow, B., S.~Szulman, and A.~J.~B. Cl{\'e}ment (1999).
\newblock Terminae: A linguistics-based tool for the building of a domain
  ontology.
\newblock In D.~Fensel and R.~Studer (Eds.), {\em Knowledge Acquisition,
  Modeling and Management}, pp.\  49--66.

\bibitem[\protect\citeauthoryear{Bojanowski, Grave, Joulin, and
  Mikolov}{Bojanowski et~al.}{2017}]{bojanowski2017enriching}
Bojanowski, P., E.~Grave, A.~Joulin, and T.~Mikolov (2017).
\newblock Enriching word vectors with subword information.
\newblock {\em Transactions of the Association for Computational
  Linguistics\/}~{\em 5}, 135--146.

\bibitem[\protect\citeauthoryear{Bouma}{Bouma}{2009}]{boumaunknownnormalized}
Bouma, G. (2009).
\newblock Normalized (pointwise) mutual information in collocation extraction.
\newblock {\em Proceedings of GSCL\/}~{\em 30}, 31--40.

\bibitem[\protect\citeauthoryear{Brown, Mann, Ryder, Subbiah, Kaplan, Dhariwal,
  Neelakantan, Shyam, Sastry, Askell, et~al.}{Brown
  et~al.}{2020}]{brown2020language}
Brown, T., B.~Mann, N.~Ryder, M.~Subbiah, J.~D. Kaplan, P.~Dhariwal,
  A.~Neelakantan, P.~Shyam, G.~Sastry, A.~Askell, et~al. (2020).
\newblock Language models are few-shot learners.
\newblock {\em Advances in neural information processing systems\/}~{\em 33},
  1877--1901.

\bibitem[\protect\citeauthoryear{Chowdhury, Gliozzo, and Trewin}{Chowdhury
  et~al.}{2018}]{chowdhury2018domain}
Chowdhury, M. F.~M., A.~M. Gliozzo, and S.~M. Trewin (2018, September~27).
\newblock Domain-specific terminology extraction by boosting frequency metrics.
\newblock US Patent App. 15/469,766.

\bibitem[\protect\citeauthoryear{Cimiano and V{\"o}lker}{Cimiano and
  V{\"o}lker}{2005}]{cimiano2005text2onto}
Cimiano, P. and J.~V{\"o}lker (2005).
\newblock text2onto.
\newblock In {\em International conference on application of natural language
  to information systems}, pp.\  227--238. Springer.

\bibitem[\protect\citeauthoryear{Cohen}{Cohen}{1960}]{cohen1960}
Cohen, J. (1960).
\newblock {A Coefficient of Agreement for Nominal Scales}.
\newblock {\em Educational and Psychological Measurement\/}~{\em 20\/}(1),
  1--10.

\bibitem[\protect\citeauthoryear{Cunningham}{Cunningham}{2002}]{Cunningham2002}
Cunningham, H. (2002).
\newblock Gate, a general architecture for text engineering.
\newblock In {\em Computers and the Humanities}, Volume~36, pp.\  223--254.

\bibitem[\protect\citeauthoryear{Daille}{Daille}{2002}]{Daille1994StudyAI}
Daille, B. (2002, 12).
\newblock Study and implementation of combined techniques for automatic
  extraction of terminology.
\newblock {\em The Balancing Act: Combining Symbolic and Statistical Approaches
  to Language\/}~{\em 1}, 1--8.

\bibitem[\protect\citeauthoryear{Devlin, Chang, Lee, and Toutanova}{Devlin
  et~al.}{2019}]{devlin-etal-2019-bert}
Devlin, J., M.-W. Chang, K.~Lee, and K.~Toutanova (2019, June).
\newblock {BERT}: Pre-training of deep bidirectional transformers for language
  understanding.
\newblock In J.~Burstein, C.~Doran, and T.~Solorio (Eds.), {\em Proceedings of
  the 2019 Conference of the North {A}merican Chapter of the Association for
  Computational Linguistics: Human Language Technologies, Volume 1 (Long and
  Short Papers)}, Minneapolis, Minnesota, pp.\  4171--4186. Association for
  Computational Linguistics.

\bibitem[\protect\citeauthoryear{Dias and Kaalep}{Dias and
  Kaalep}{2003}]{Dias2003AutomaticEO}
Dias, G. and H.-J. Kaalep (2003).
\newblock Automatic extraction of multiword units for estonian : Phrasal verbs.
\newblock In {\em Languages in Development}, pp.\  41:81–91.

\bibitem[\protect\citeauthoryear{Drouin}{Drouin}{2003}]{Drouin2003}
Drouin, P. (2003).
\newblock Term extraction using non technical corpora as point of leverage.
\newblock In n.~Terminology, vol.~9 (Ed.), {\em John Benjamins Publishing
  Company: Amsterdam/Philadelphia}, pp.\  99--115.

\bibitem[\protect\citeauthoryear{Gillam, Tariq, and Ahmad}{Gillam
  et~al.}{2005}]{Gillam05terminologyand}
Gillam, L., M.~Tariq, and K.~Ahmad (2005).
\newblock Terminology and the construction of ontology.
\newblock {\em TERMINOLOGY\/}~{\em 11}, 55--81.

\bibitem[\protect\citeauthoryear{Goutte and Gaussier}{Goutte and
  Gaussier}{2005}]{Gaussier}
Goutte, C. and E.~Gaussier (2005).
\newblock { A Probabilistic Interpretation of Precision, Recall and F-Score,
  with Implication for Evaluation}.
\newblock {\em ECIR 2005\/}~{\em 1}, 345--359.

\bibitem[\protect\citeauthoryear{Harris}{Harris}{1968}]{harris68}
Harris, Z.~S. (1968).
\newblock {\em Mathematical Structures of Language}.
\newblock New York: Wiley.

\bibitem[\protect\citeauthoryear{Hearst}{Hearst}{1992}]{hearst1992}
Hearst, M.~A. (1992).
\newblock Automatic acquisition of hyponyms from large text corpora.
\newblock In {\em {COLING} 1992}, pp.\  1--7.

\bibitem[\protect\citeauthoryear{Hunter, Louradour, Rennard, Harrando, Shang,
  and Lorr{\'e}}{Hunter et~al.}{2023}]{hunter2023claire}
Hunter, J., J.~Louradour, V.~Rennard, I.~Harrando, G.~Shang, and J.-P.
  Lorr{\'e} (2023).
\newblock The claire french dialogue dataset.
\newblock {\em arXiv preprint arXiv:2311.16840\/}~{\em 1}, 1--11.

\bibitem[\protect\citeauthoryear{Kessler, B\'{e}chet, and Berio}{Kessler
  et~al.}{2019}]{Kessler2019}
Kessler, R., N.~B\'{e}chet, and G.~Berio (2019).
\newblock Extraction of terminology in the field of construction.
\newblock In {\em Proceedings of DDP 2019}, pp.\  22--26.

\bibitem[\protect\citeauthoryear{Kluegl, Toepfer, Beck, Fette, and
  Puppe}{Kluegl et~al.}{2016}]{kluegl_toepfer_beck_fette_puppe_2016}
Kluegl, P., M.~Toepfer, P.-D. Beck, G.~Fette, and F.~Puppe (2016).
\newblock Uima ruta: Rapid development of rule-based information extraction
  applications.
\newblock {\em Natural Language Engineering\/}~{\em 22\/}(1), 1–40.

\bibitem[\protect\citeauthoryear{Lang, Schneider, and Suchowolec}{Lang
  et~al.}{2018}]{lang2018extracting}
Lang, C., R.~Schneider, and K.~Suchowolec (2018).
\newblock Extracting specialized terminology from linguistic corpora.
\newblock {\em GRAMMAR AND CORPORA\/}~{\em 1}, 425.

\bibitem[\protect\citeauthoryear{Lefever, Macken, and Hoste}{Lefever
  et~al.}{2009}]{Lefever2009}
Lefever, E., L.~Macken, and V.~Hoste (2009).
\newblock Language-independent bilingual terminology extraction from a
  multilingual parallel corpus.
\newblock In {\em Proceedings of the 12th Conference of the European Chapter of
  the Association for Computational Linguistics}, EACL '09, Stroudsburg, PA,
  USA, pp.\  496--504. Association for Computational Linguistics.

\bibitem[\protect\citeauthoryear{Lenci and Benotto}{Lenci and
  Benotto}{2012}]{lenci2012}
Lenci, A. and G.~Benotto (2012, 7-8 June).
\newblock Identifying hypernyms in distributional semantic spaces.
\newblock In {\em *Proceedings of ({S}em{E}val 2012)}, Montr{\'e}al, Canada,
  pp.\  75--79. Association for Computational Linguistics.

\bibitem[\protect\citeauthoryear{Levy, Remus, Biemann, and Dagan}{Levy
  et~al.}{2015}]{levy2015}
Levy, O., S.~Remus, C.~Biemann, and I.~Dagan (2015, May{--}June).
\newblock Do supervised distributional methods really learn lexical inference
  relations?
\newblock In {\em Proceedings of NACL 2015}, Denver, Colorado, pp.\  970--976.

\bibitem[\protect\citeauthoryear{Lin}{Lin}{1998}]{Lin1998}
Lin, D. (1998).
\newblock An information-theoretic definition of similarity.
\newblock In {\em Proceedings of ICML '98}, San Francisco, CA, USA, pp.\
  296--304. Morgan Kaufmann Publishers Inc.

\bibitem[\protect\citeauthoryear{Liu and Singh}{Liu and Singh}{2004}]{Liu2004}
Liu, H. and P.~Singh (2004, October).
\newblock Conceptnet \&mdash; a practical commonsense reasoning tool-kit.
\newblock {\em BT Technology Journal\/}~{\em 22\/}(4), 211--226.

\bibitem[\protect\citeauthoryear{Liu, Ott, Goyal, Du, Joshi, Chen, Levy, Lewis,
  Zettlemoyer, and Stoyanov}{Liu et~al.}{2019}]{liu2019roberta}
Liu, Y., M.~Ott, N.~Goyal, J.~Du, M.~Joshi, D.~Chen, O.~Levy, M.~Lewis,
  L.~Zettlemoyer, and V.~Stoyanov (2019).
\newblock Roberta: A robustly optimized bert pretraining approach.

\bibitem[\protect\citeauthoryear{Luu, Tay, Hui, and Ng}{Luu
  et~al.}{2016}]{luu2016}
Luu, A.~T., Y.~Tay, S.~C. Hui, and S.~K. Ng (2016, November).
\newblock Learning term embeddings for taxonomic relation identification using
  dynamic weighting neural network.
\newblock In {\em Proceedings of the 2016 Conference on EMNLP}, Austin, Texas,
  pp.\  403--413.

\bibitem[\protect\citeauthoryear{Ma, Wang, Bao, He, and Zheng}{Ma
  et~al.}{2023}]{ma-etal-2023-kepl}
Ma, N., D.~Wang, H.~Bao, L.~He, and S.~Zheng (2023, December).
\newblock {KEPL}: Knowledge enhanced prompt learning for {C}hinese
  hypernym-hyponym extraction.
\newblock In H.~Bouamor, J.~Pino, and K.~Bali (Eds.), {\em Proceedings of the
  2023 Conference on Empirical Methods in Natural Language Processing},
  Singapore, pp.\  5858--5867. Association for Computational Linguistics.

\bibitem[\protect\citeauthoryear{Macken, Lefever, and Hoste}{Macken
  et~al.}{2013}]{Macken2013TExSISBT}
Macken, L., E.~Lefever, and V.~Hoste (2013).
\newblock Texsis: bilingual terminology extraction from parallel corpora using
  chunk-based alignment.
\newblock {\em Terminology\/}~{\em 19\/}(1), 1--30.

\bibitem[\protect\citeauthoryear{Markchom, Liang, and Chen}{Markchom
  et~al.}{2022}]{markchom-etal-2022-uor}
Markchom, T., H.~Liang, and J.~Chen (2022, July).
\newblock {U}o{R}-{NCL} at {S}em{E}val-2022 task 3: Fine-tuning the
  {BERT}-based models for validating taxonomic relations.
\newblock In G.~Emerson, N.~Schluter, G.~Stanovsky, R.~Kumar, A.~Palmer,
  N.~Schneider, S.~Singh, and S.~Ratan (Eds.), {\em Proceedings of the 16th
  International Workshop on Semantic Evaluation (SemEval-2022)}, Seattle,
  United States, pp.\  260--265. Association for Computational Linguistics.

\bibitem[\protect\citeauthoryear{Martin, Muller, Ortiz~Su{\'a}rez, Dupont,
  Romary, de~la Clergerie, Seddah, and Sagot}{Martin
  et~al.}{2020}]{martin-etal-2020-camembert}
Martin, L., B.~Muller, P.~J. Ortiz~Su{\'a}rez, Y.~Dupont, L.~Romary,
  {\'E}.~de~la Clergerie, D.~Seddah, and B.~Sagot (2020, July).
\newblock {C}amem{BERT}: a tasty {F}rench language model.
\newblock In {\em Proceedings of the 58th Annual Meeting of the Association for
  Computational Linguistics}, Online, pp.\  7203--7219. Association for
  Computational Linguistics.

\bibitem[\protect\citeauthoryear{McHugh}{McHugh}{2012}]{McHugh2012InterraterRT}
McHugh, M.~L. (2012).
\newblock Interrater reliability: the kappa statistic.
\newblock In {\em Biochemia medica}, pp.\  1--7.

\bibitem[\protect\citeauthoryear{Meignier and Merlin}{Meignier and
  Merlin}{2010}]{Meignier10liumspkdiarization}
Meignier, S. and T.~Merlin (2010).
\newblock Lium spkdiarization: an open source toolkit for diarization.
\newblock In {\em in CMU SPUD Workshop}.

\bibitem[\protect\citeauthoryear{Meyers, He, Glass, Ortega, Liao, Grieve-Smith,
  Grishman, and Babko-Malaya}{Meyers et~al.}{2018}]{Meyers2018}
Meyers, A.~L., Y.~He, Z.~Glass, J.~Ortega, S.~Liao, A.~Grieve-Smith,
  R.~Grishman, and O.~Babko-Malaya (2018).
\newblock The termolator: Terminology recognition based on chunking,
  statistical and search-based scores.
\newblock {\em Frontiers in Research Metrics and Analytics\/}~{\em 3}, 1--14.

\bibitem[\protect\citeauthoryear{Mikolov, Grave, Bojanowski, Puhrsch, and
  Joulin}{Mikolov et~al.}{2018}]{mikolov2018advances}
Mikolov, T., E.~Grave, P.~Bojanowski, C.~Puhrsch, and A.~Joulin (2018).
\newblock Advances in pre-training distributed word representations.
\newblock In {\em Proceedings of the International Conference on Language
  Resources and Evaluation (LREC 2018)}, pp.\  1--4.

\bibitem[\protect\citeauthoryear{Mikolov, Sutskever, Chen, Corrado, and
  Dean}{Mikolov et~al.}{2013}]{Mikolov2013}
Mikolov, T., I.~Sutskever, K.~Chen, G.~Corrado, and J.~Dean (2013).
\newblock Distributed representations of words and phrases and their
  compositionality.
\newblock In {\em Proceedings of NIPS'13}, USA, pp.\  3111--3119. Curran
  Associates Inc.

\bibitem[\protect\citeauthoryear{Mirkin, Dagan, and Geffet}{Mirkin
  et~al.}{2006}]{mirkin2006}
Mirkin, S., I.~Dagan, and M.~Geffet (2006, July).
\newblock Integrating pattern-based and distributional similarity methods for
  lexical entailment acquisition.
\newblock In {\em Proceedings of the {COLING}/{ACL} 2006}, Sydney, Australia,
  pp.\  579--586. ACL.

\bibitem[\protect\citeauthoryear{Panchenko, Faralli, Ruppert, Remus, Naets,
  Fairon, Ponzetto, and Biemann}{Panchenko et~al.}{2016}]{taxi2016}
Panchenko, A., S.~Faralli, E.~Ruppert, S.~Remus, H.~Naets, C.~Fairon, S.~P.
  Ponzetto, and C.~Biemann (2016).
\newblock {TAXI} at semeval-2016 task 13: a taxonomy induction method based on
  lexico-syntactic patterns, substrings and focused crawling.
\newblock In {\em Proceedings of the 10th International Workshop on Semantic
  Evaluation, SemEval@NAACL-HLT 2016, San Diego, CA, USA, June 16-17, 2016},
  pp.\  1320--1327.

\bibitem[\protect\citeauthoryear{Pauwels and Terkaj}{Pauwels and
  Terkaj}{2016}]{Pauwels2016}
Pauwels, P. and W.~Terkaj (2016, 03).
\newblock Express to owl for construction industry: Towards a recommendable and
  usable ifcowl ontology.
\newblock {\em Automation in Construction\/}~{\em 63}, 100--133.

\bibitem[\protect\citeauthoryear{Pennington, Socher, and Manning}{Pennington
  et~al.}{2014}]{Pennington2014}
Pennington, J., R.~Socher, and C.~D. Manning (2014).
\newblock Glove: Global vectors for word representation.
\newblock In {\em Proceedings of the 2014 conference on empirical methods in
  natural language processing (EMNLP)}, pp.\  1532--1543.

\bibitem[\protect\citeauthoryear{Roche and Kodratoff}{Roche and
  Kodratoff}{2004}]{Roche06}
Roche, M. and Y.~Kodratoff (2004).
\newblock Exit: Un système itératif pour l’extraction de la terminologie du
  domaine à partir de corpus spécialisés.
\newblock In {\em Proceedings of JADT 4}, pp.\  946--956.

\bibitem[\protect\citeauthoryear{Roller, Erk, and Boleda}{Roller
  et~al.}{2014}]{roller2014}
Roller, S., K.~Erk, and G.~Boleda (2014, August).
\newblock Inclusive yet selective: Supervised distributional hypernymy
  detection.
\newblock In {\em Proceedings of {COLING} 2014}, Dublin, Ireland, pp.\
  1025--1036. Dublin City University and ACL.

\bibitem[\protect\citeauthoryear{Salton and McGill}{Salton and
  McGill}{1986}]{Salton1986}
Salton, G. and M.~J. McGill (1986).
\newblock {\em Introduction to Modern Information Retrieval}.
\newblock New York, NY, USA: McGraw-Hill, Inc.

\bibitem[\protect\citeauthoryear{Santus, Yung, Lenci, and Huang}{Santus
  et~al.}{2015}]{santus2015}
Santus, E., F.~Yung, A.~Lenci, and C.-R. Huang (2015, July).
\newblock {EVAL}ution 1.0: an evolving semantic dataset for training and
  evaluation of distributional semantic models.
\newblock In {\em Proceedings of the 4th Workshop on LDL}, pp.\  64--69.

\bibitem[\protect\citeauthoryear{Schmid}{Schmid}{1994}]{schmid1994probabilistic}
Schmid, H. (1994).
\newblock Probabilistic part-of-speech tagging using decision trees.

\bibitem[\protect\citeauthoryear{Shwartz, Santus, and Schlechtweg}{Shwartz
  et~al.}{2016}]{Shwartz2016}
Shwartz, V., E.~Santus, and D.~Schlechtweg (2016).
\newblock Hypernyms under siege: Linguistically-motivated artillery for
  hypernymy detection.
\newblock {\em CoRR\/}~{\em abs/1612.04460}, 425--435.

\bibitem[\protect\citeauthoryear{Touvron, Martin, Stone, Albert, Almahairi,
  Babaei, Bashlykov, Batra, Bhargava, Bhosale, et~al.}{Touvron
  et~al.}{2023}]{touvron2023llama}
Touvron, H., L.~Martin, K.~Stone, P.~Albert, A.~Almahairi, Y.~Babaei,
  N.~Bashlykov, S.~Batra, P.~Bhargava, S.~Bhosale, et~al. (2023).
\newblock Llama 2: Open foundation and fine-tuned chat models.
\newblock {\em arXiv preprint arXiv:2307.09288\/}~{\em 1}, 510--521.

\bibitem[\protect\citeauthoryear{Vetter, Segiet, and Lennermann}{Vetter
  et~al.}{2022}]{vetter-etal-2022-kamikla}
Vetter, K., M.~Segiet, and K.~Lennermann (2022, July).
\newblock {K}a{M}i{K}la at {S}em{E}val-2022 task 3: {A}l{BERT}o, {BERT}, and
  {C}amem{BERT}{---}{B}e(r)tween taxonomy detection and prediction.
\newblock In G.~Emerson, N.~Schluter, G.~Stanovsky, R.~Kumar, A.~Palmer,
  N.~Schneider, S.~Singh, and S.~Ratan (Eds.), {\em Proceedings of
  SemEval-202}, Seattle, United States, pp.\  282--290. ACL.

\bibitem[\protect\citeauthoryear{Weeds, Clarke, Reffin, Weir, and Keller}{Weeds
  et~al.}{2014}]{weeds2014}
Weeds, J., D.~Clarke, J.~Reffin, D.~Weir, and B.~Keller (2014, August).
\newblock Learning to distinguish hypernyms and co-hyponyms.
\newblock In {\em Proceedings of {COLING} 2014}, Dublin, Ireland, pp.\
  2249--2259.

\bibitem[\protect\citeauthoryear{Weeds, Weir, and McCarthy}{Weeds
  et~al.}{2004}]{weeds2004}
Weeds, J., D.~Weir, and D.~McCarthy (2004, aug 23{--}aug 27).
\newblock Characterising measures of lexical distributional similarity.
\newblock In {\em Proceedings of {COLING} 2004}, pp.\  1015--1021.

\bibitem[\protect\citeauthoryear{Wohlgenannt and Minic}{Wohlgenannt and
  Minic}{2016}]{wohlgenannt2016using}
Wohlgenannt, G. and F.~Minic (2016).
\newblock Using word2vec to build a simple ontology learning system.
\newblock In {\em International Semantic Web Conference (Posters \& Demos)}.

\end{thebibliography}
\bibliographystyle{chicago}

\end{document}